\documentclass[draftclasofoot, onecolumn]{IEEEtran}
\IEEEoverridecommandlockouts

\usepackage[noadjust]{cite}
\usepackage[bookmarks=true,breaklinks=true,colorlinks,linkcolor=red,citecolor=blue,urlcolor=BlueViolet]{hyperref}
\usepackage{multirow}
\usepackage{amsmath,amssymb,amsfonts}
\usepackage[ruled,vlined]{algorithm2e}
\newcommand{\mycommentstyle}[1]{\color[HTML]{0671b9}{#1}}
\SetKwComment{Comment}{\mycommentstyle{// }}{}
\usepackage{graphicx}
\usepackage{threeparttable}
\usepackage{subfigure}
\usepackage{textcomp}
\usepackage{xcolor}
\def\BibTeX{{\rm B\kern-.05em{\sc i\kern-.025em b}\kern-.08em
    T\kern-.1667em\lower.7ex\hbox{E}\kern-.125emX}}
\usepackage{booktabs}
\usepackage{tablefootnote}
\usepackage{blindtext}
\usepackage{float}
\usepackage{algorithmic}
\usepackage{textcomp}
\usepackage{booktabs}

\begin{document}

\title{A Causal Convolutional Low-rank Representation Model for Imputation of Water Quality Data}

\author{\IEEEauthorblockN{Xin Liao}\\
	\IEEEauthorblockA{\textit{College of Computer and Information Science}
		\textit{Southwest University}
		Chongqing, China \\
		lxchat26@gmail.com}
\\
	\IEEEauthorblockN{Bing Yang}\\
	\IEEEauthorblockA{\textit{School of Environment and Ecology} 
		\textit{Chongqing University}\\
		Chongqing, China \\
		cq\_yangbing@163.com}
\\
	\IEEEauthorblockN{Tan Dongli}\\
	\IEEEauthorblockA{\textit{Chongqing Eco-Environment Monitoring Center} 
		Chongqing, China \\
		406865661@qq.com}
\\
	\IEEEauthorblockN{Cai Yu*}\\
	\IEEEauthorblockA{\textit{Chongqing Eco-Environment Monitoring Center}
		Chongqing, China \\
		cyscut@foxmail.com}
}

\maketitle

\begin{abstract}
The monitoring of water quality is a crucial part of environmental protection, and a large number of monitors are widely deployed to monitor water quality. Due to unavoidable factors such as data acquisition breakdowns, sensors and communication failures, water quality monitoring data suffers from missing values over time, resulting in High-Dimensional and Sparse (HDS) Water Quality Data (WQD). The simple and rough filling of the missing values leads to inaccurate results and affects the implementation of relevant measures. Therefore, this paper proposes a Causal convolutional Low-rank Representation (CLR) model for imputing missing WQD to improve the completeness of the WQD, which employs a two-fold idea: a) applying causal convolutional operation to consider the temporal dependence of the low-rank representation, thus incorporating temporal information to improve the imputation accuracy; and b) implementing a hyperparameters adaptation scheme to automatically adjust the best hyperparameters during model training, thereby reducing the tedious manual adjustment of hyper-parameters. Experimental studies on three real-world water quality datasets demonstrate that the proposed CLR model is superior to some of the existing state-of-the-art imputation models in terms of imputation accuracy and time cost, as well as indicating that the proposed model provides more reliable decision support for environmental monitoring.
\end{abstract}


\section{Introduction}
Water quality monitoring plays a key role in environmental protection, public health and sustainable development. With the advancement of industrialization and intensification of human activities, water resource management becomes more and more important, and accurate Water Quality Data (WQD) provide a scientific basis for water resource management, pollution prevention and ecological protection~\cite{zeng2016cost}. By collecting WQD regularly, water quality trends are effectively identified and timely measures are taken to ensure water quality. However, the data recorded by widely deployed monitors are missing data due to equipment and transmission failures during data collection, and the High-Dimensional and Sparse (HDS)~\cite{r1,r2,r3,r4,r5,r6,r7} WQD seriously affects data completeness, reliability, and hinders the accurate assessment of water quality monitoring. Therefore, how to precisely impute HDS data to provide subsequent analysis becomes a tricky task~\cite{r21,r8,r9}.

In recent years, researchers propose a variety of missing data imputation models to effectively impute missing values in data, such as K-Nearest Neighbors (KNN) imputation models, decision trees methods, Support Vector Machines (SVM) imputation, and deep learning approaches. Specifically, Murti \textit{et al.}~\cite{8987530} adopt the KNN imputation to capture complex relationships in data and thus impute missing data, but it is difficult to effectively identify the true neighbors for high-dimensional data to provide accurate imputation. Rahman \textit{et al.}~\cite{rahman2013missing} utilize decision trees to model complex relationships between features and capture nonlinear relationships to offer accurate imputation. However, this approach is easily suffer from the risk of overfitting. Brand \textit{et al.}~\cite{brand2021multi} apply SVM imputation to inscribe high-dimensional nonlinear relationships in the data and maintain robustness by regularizing the parameters, however it suffers from high computational complexity for large datasets. Samad \textit{et al.}\cite{samad2022missing} propose an integrated learning-based imputation method that utilizes Deep Neural Networks (DNN) to replace linear regressors to improve the accuracy of imputing missing data. Unfortunately, deep learning approaches require considerable computational and time cost due to the large number of parameters computed~\cite{r14,r16,r30}.

\begin{figure*}[t]
	\centering
	\includegraphics[width=0.8\linewidth]{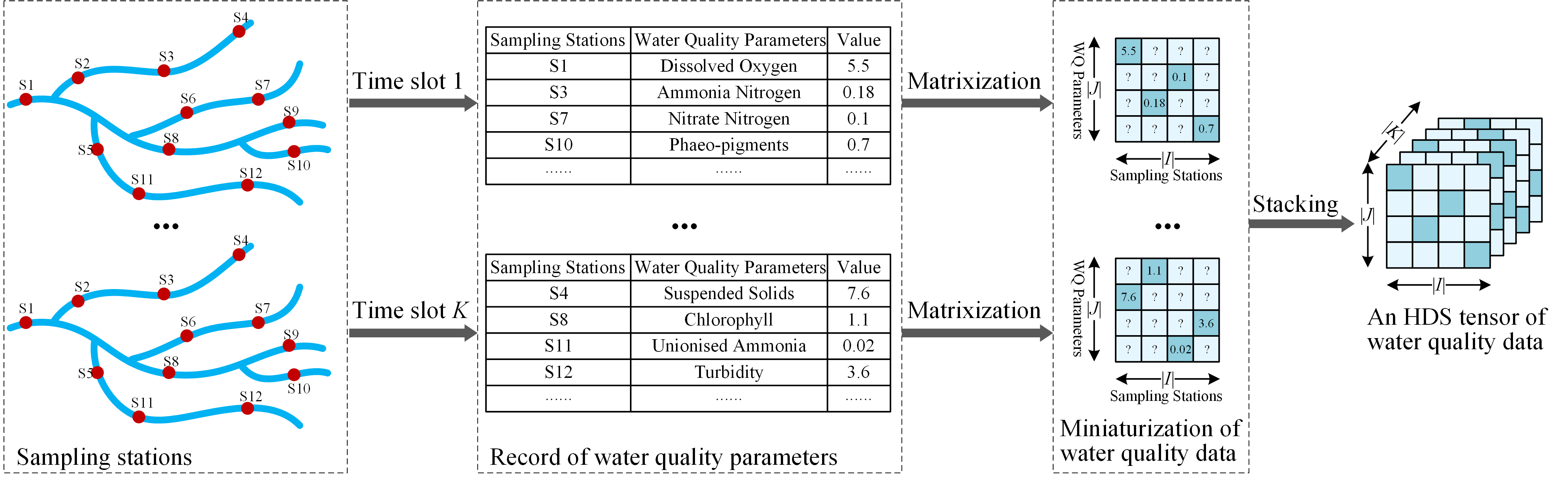}
	\caption{An illustration of water quality data constructed as an HDS tensor.}
	\label{wqd}
\end{figure*}

Previous studies demonstrate that the Tensor Low-rank Representation (TLR) model can effectively deal with HDS data by modeling the features of the data via compact vector representations in the low-rank space to impute missing values in the HDS tensor~\cite{r22, r10, r78, r12}. On the other hand, the WQD can be represented as an HDS tensor presented in Fig.~\ref{wqd}. Specifically, widely distributed sampling stations recorded monitoring data for each time slot, i.e., the monitoring values of the sampling stations for water quality parameters. By matrixizing the WQD for each time slot, a series of matrices arranged by time are obtained. Finally, an HDS tensor is constructed by stacking the above matrices, and its an elements denotes the monitoring value of the sampling station $ i \in I $ for the water quality parameter $ j \in J $ at the time slot $ k \in K $. Therefore, we propose a Causal convolutional Low-rank Representation (CLR) model for imputing the WQD, which employs the causal convolutional operation and a Particle Swarm Optimization (PSO)-based~\cite{r20, r46, r52, r70} hyperparameters adaptation scheme to offer high imputation accuracy and low time cost~\cite{r17, r18, r19, r24, r31, r32}. The main contributions of this paper are listed as follows:
\begin{itemize}
	\item A causal convolutional operator. It considers the temporal correlation of low-rank representations and efficiently models temporal features via causal characterization on the time, thereby providing accurate imputation accuracy.
	\item A PSO-based hyperparameters adaptation scheme. It avoids the tedious hyperparameters tuning process and adaptively adjusts the best hyperparameters during model training to reduce the model's time overhead.
	\item An empirical study on three datasets. It demonstrates that the proposed CLR model outperforms several state-of-the-art imputation models in terms of imputation accuracy and time cost.
\end{itemize}

The remaining Sections are organized as follows: Section~\ref{back} presents the background, Section~\ref{model} introduces the proposed CLR model, Section~\ref{comparisons} and~\ref{conclusions} provide experimental comparisons and conclusions, respectively.

\section{Background}\label{back}
\subsection{Notation system}

\begin{table}[t]
	\caption{The description for the notation}\label{t1}
	\centering
	\begin{tabular}{@{}cl@{}}
		\toprule
		Notation & Description \\ \midrule
		$ I $ & A Set of sampling stations. \\
		$ J $ & A set of water quality parameters. \\
		$ K $ & A set of time slots. \\
		$\textbf{X} $ & An HDS tensor constructed via the WQD. \\
		$\tilde{\textbf{X}} $ & The low-rank approximation tensor of \textbf{X}.  \\ 
		$\textbf{D} $ & rank-one tensors for $\tilde{\textbf{X}}$. \\
		$ x_{ijk}, \tilde{x}_{ijk}, d_{ijk} $  & A single element of $ \textbf{X}, \tilde{\textbf{X}} $, and $ \textbf{R} $. \\
		$ R $ & The number of rank-one tensors. \\
		$ \mathrm{S}, \mathrm{U}, \mathrm{V} $ & Three feature matrices correspond to $ I, J, K $. \\
		$ \textbf{\textit{s}}_{r}, \textbf{\textit{u}}_{r}, \textbf{\textit{v}}_{r} $ & The $ r $-th column vectors of S, U, and V. \\
		$ s_{ir}, u_{jr}, v_{kr} $ & An single element of S, U, and V. \\
		$ \mathrm{W} $ & The weights of causal convolutional operator. \\
		$ w_{cr} $ & An single weight of W. \\
		$ \textbf{\textit{a}}, \textbf{\textit{e}}, \textbf{\textit{o}} $ & Three linear bias vectors. \\
		$ a_{i}, e_{j}, o_{k} $ & The single element of $ \textbf{\textit{a}} $, $ \textbf{\textit{e}} $, and $ \textbf{\textit{o}} $. \\ \bottomrule
	\end{tabular}
	\label{symbol}
\end{table}

The notations employed in this paper are summarized in Table~\ref{t1}. The tensor is denoted in uppercase bold, the matrix is denoted in uppercase, the vector is denoted in lowercase bold italic, and the scalar is denoted in lowercase italic.

\subsection{Problem formulation}

As presented in Fig.~\ref{wqd}, the WQD can be represented by an HDS tensor, which is defined as:

\textbf{Definition 1.} (HDS tensor). For a given tensor $ \textbf{X} \in \mathbb{R}^{|I|\times|J|\times|K|} $, let $ \Omega $ and $ \Upsilon $ denote the observed and unobserved element sets in the tensor, respectively. If $ |\Upsilon|  \gg   |\Omega| $, the tensor $ \textbf{X} $ is an HDS tensor, where a element $ x_{ijk} $ denotes a quantized value determined by the subscript $ (i,j,k) $.

Following the CPD principle~\cite{r36,r37,r38,r39}, the HDS tensor $ \textbf{X} $ is decomposed into three low-rank feature matrices $ {\rm{S}} \in \mathbb{R}^{|I| \times R} $, $ {\rm{U}} \in \mathbb{R}^{|J| \times R} $, and $ {\rm{V}} \in \mathbb{R}^{|K| \times R} $ to represent the sampling station features, water quality parameter features, and temporal features, respectively. $ R $ denotes the feature dimension also the number of rank-one tensors, i.e., the low-rank approximation tensor $ \tilde{\textbf{X}} $ is constructed by $ R $ rank-one tensors~\cite{r25,r26,r29,r33} as:
\begin{equation}\label{e1}
	\tilde{\textbf{X}} = \sum\limits_{r = 1}^R {{\textbf{D}_r}} ,
\end{equation}
where the rank-one tensor $ \textbf{D}_r $  is obtained by the outer product for the $ r $-th column vectors of the three feature matrices, i.e. $ \mathbf{D}_r= \textbf{\textit{s}}_r \circ \textbf{\textit{u}}_r \circ  \textbf{\textit{v}}_r $. Therefore, the low-rank approximation tensor $ \tilde{\textbf{X}} $ is formulated as:
\begin{equation}
	\tilde{\textbf{X}} = \sum\limits_{r = 1}^R {\textbf{\textit{s}}_r \circ \textbf{\textit{u}}_r \circ  \textbf{\textit{v}}_r}.
\end{equation}
With a finer-grained representation, a single element $ \tilde{x}_{ijk} $ in $ \tilde{\textbf{X}} $ is obtained as:
\begin{equation}\label{e3}
	{\tilde{x}_{ijk}} = \sum\limits_{r = 1}^R {d_{ijk}^{(r)}}  = \sum\limits_{r = 1}^R {{s_{ir}}{u_{jr}}{v_{kr}}} .
\end{equation}

\begin{figure}[t]
	\centering
	\includegraphics[width=0.8\linewidth]{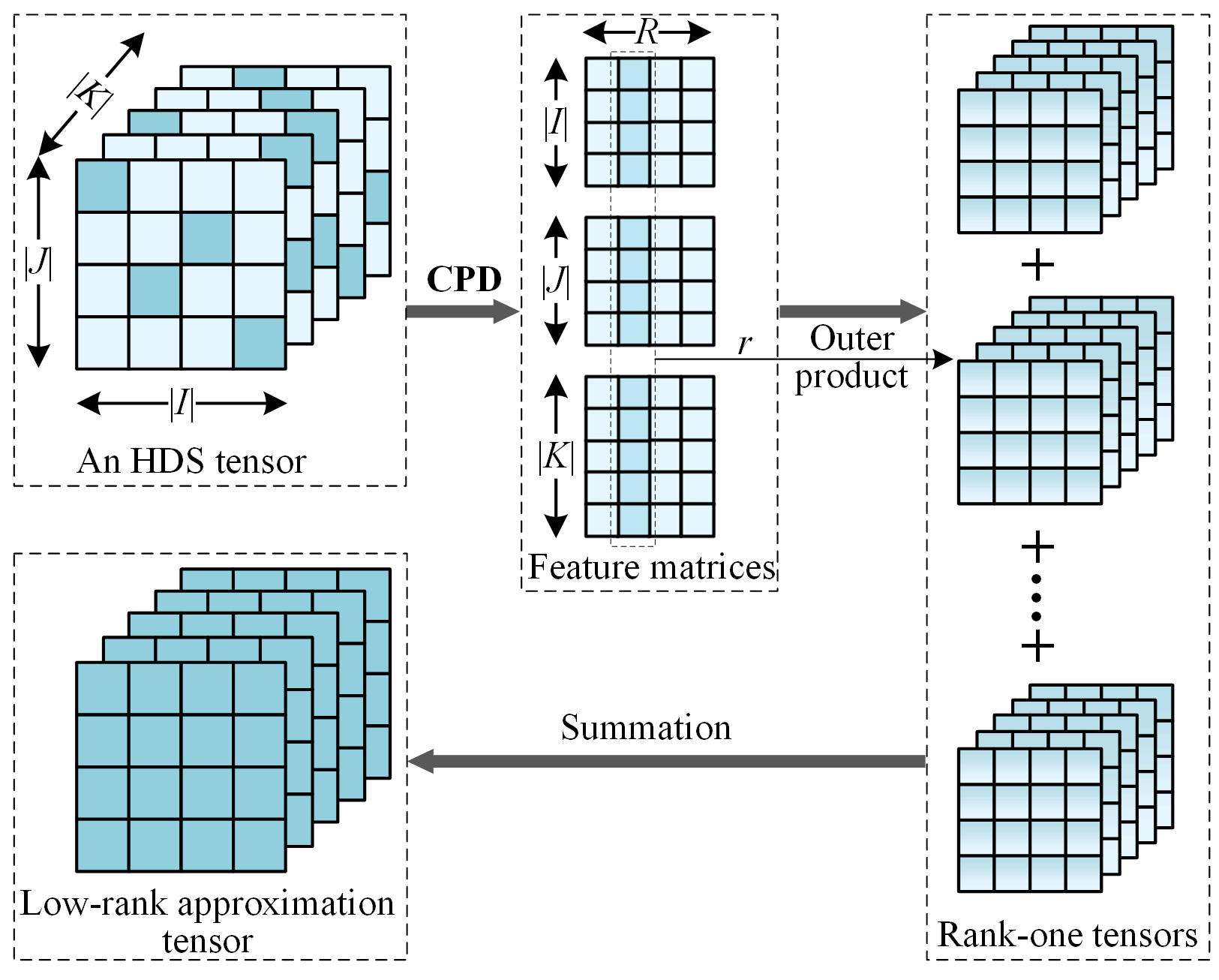}
	\caption{An illustrative example of a tensor low-rank representation.}
	\label{lft}
\end{figure}

With (\ref{e1})-(\ref{e3}), an HDS tensor is low-rank represented by three feature matrices as illustrated in Fig.~\ref{lft}.

As indicated by previous studies~\cite{r41,r42,r43,r44,r45}, the low-rank representation models incorporating linear bias vectors are effective in suppressing fluctuations during model training process for the HDS data. Therefore, the element $ \tilde{x}_{ijk} $ is reformulated as:
\begin{equation}
	{\tilde{x}_{ijk}} = \sum\limits_{r = 1}^R {{s_{ir}}{u_{jr}}{v_{kr}}} + a_i + e_j + o_k,
\end{equation}
where $ a_i $, $e_j $, and $ o_k $ are the elements of $ \textbf{\textit{a}} \in \mathbb{R}^{|I|}, \textbf{\textit{e}} \in \mathbb{R}^{|J|}  $, and $ \textbf{\textit{o}} \in \mathbb{R}^{|K|} $ with subscripts $ i, j, k $, respectively. To achieve the desired feature matrices and bias vectors, an objective function employing Euclidean distances~\cite{r48,r49,r50,r51,r54,r55} is commonly constructed to evaluate the difference between the original tensor and the approximated tensor. On the other hand, evaluating the gap for above two tensors on the observed element set $ \Omega $ can effectively deal with data imbalance in the HDS tensor. Therefore, the objective function $ \varepsilon $ is constructed as:
\begin{equation}
	\begin{array}{l}
		\varepsilon  = \frac{1}{2}\sum\limits_{x_{ijk} \in \Omega}  {{{\left( {{x_{ijk}} - \sum\limits_{r = 1}^R {{s_{ir}}{u_{jr}}{v_{kr}}}  - {a_i} - {e_j} - {o_k}} \right)}^2}}.
	\end{array}
\end{equation}
To avoid overfitting of the model training, the $ L_2 $ regularization term is incorporated into the objective function as:
\begin{equation}
	\begin{array}{*{20}{l}}
		{\varepsilon  = \frac{1}{2}\sum\limits_{{x_{ijk}} \in \Omega } {{{\left( {{x_{ijk}} - \sum\limits_{r = 1}^R {{s_{ir}}{u_{jr}}{v_{kr}}}  - {a_i} - {e_j} - {o_k}} \right)}^2}} }\\
		{ + \frac{\lambda }{2}\sum\limits_{{x_{ijk}} \in \Omega } {\left( {\sum\limits_{r = 1}^R {s_{ir}^2 + u_{jr}^2 + v_{kr}^2}  + a_i^2 + e_j^2 + o_k^2} \right)} ,}
	\end{array}
\end{equation}
where $ \lambda $ denote the regularization constants.

\section{The CLR model}\label{model}
\subsection{Objective function}
\begin{figure}[t]
	\centering
	\includegraphics[width=0.8\linewidth]{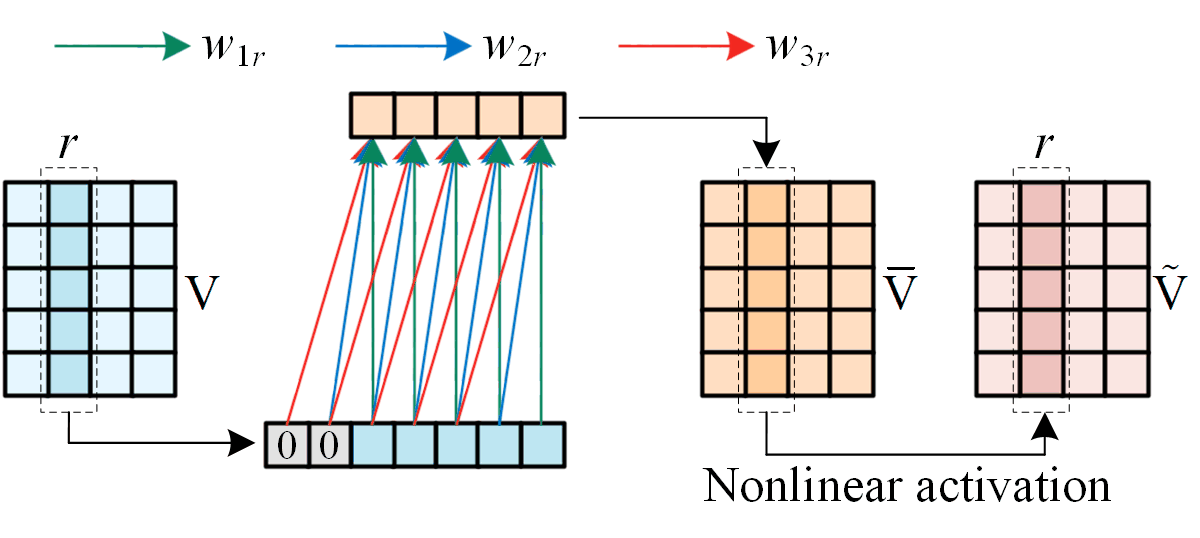}
	\caption{An illustration of the causal convolutional transform for the time feature matrix with a 3$ \times $1 convolution kernel size.}
	\label{cco}
\end{figure}

As previously discussed, the WQD is an HDS tensor that contains time-varying information~\cite{r80, r81,r82,r83,r84,r85,r86}, i.e., the data in the $ k $-th time slot influences the data in the $ (k+1) $-th time slot. Therefore, it is reasonable to implement the fusion of temporal information~\cite{r56,r58,r59} on the current data by adopting the causal convolutional operator. Specifically, we employ local temporal information incorporation for the $ k $-th time feature vector in the low-rank representation space as shown in Fig.~\ref{cco}. For the $ r $-th column of the time feature vector, it is formulated as:
\begin{equation}
	{\bar{\textbf{\textit{v}}}_r} = {\textbf{\textit{v}}_r} * {\textbf{\textit{w}}_r},
\end{equation}
where $ * $ denotes the convolutional operator and $ {\textbf{\textit{w}}_r} \in \mathbb{R}^{C} $ denotes the weight vector in the $ r $-th channel. For an element $ \bar{v}_{kr} $ in the feature vector, it is fine-grained formulated as:
\begin{equation}
	{\bar v_{kr}} = \sum\limits_{c = 1}^C {{v_{\left( {k - c + 1} \right)r}}{w_{cr}}}.
\end{equation}
Note that we employ zero padding to keep the length of the transformed vector constant. In addition, a sigmoid activation function is employed to capture nonlinear features, namely:
\begin{equation}
	{\tilde v_{kr}} = \phi \left( {{{\bar v}_{kr}}} \right),
\end{equation}
where $ \phi $ denotes the Sigmoid activation function. By implementing the nonlinear mapping to the approximation, the $ \tilde{x}_{ijk} $ is reconstructed as:
\begin{equation}
	{\tilde{x}_{ijk}} = \phi \left( \sum\limits_{r = 1}^R {{s_{ir}}{u_{jr}}{\tilde{v}_{kr}}} + a_i + e_j + o_k\right).
\end{equation}

Hence, the objective function of the proposed CLR model is constructed as:
\begin{equation}
	\begin{array}{l}
		\varepsilon  = \frac{1}{2}\sum\limits_{{x_{ijk}} \in \Omega } {{{\left( {{x_{ijk}} - \phi \left( {\sum\limits_{r = 1}^R {{s_{ir}}{u_{jr}}{{\tilde v}_{kr}}}  + {a_i} + {e_j} + {o_k}} \right)} \right)}^2}} \\
		+ \frac{\lambda }{2}\sum\limits_{{x_{ijk}} \in \Omega } {\left( {\sum\limits_{r = 1}^R {s_{ir}^2 + u_{jr}^2 + v_{kr}^2 + \sum\limits_{c = 1}^C {w_{cr}^2} }  + a_i^2 + e_j^2 + o_k^2} \right)} 
	\end{array}
\end{equation}

\subsection{Parameters learning}
In order to learn the required model parameters, we use the standard Stochastic Gradient Descent (SGD)~\cite{r63,r64} algorithm to update the feature matrices, bias vectors, and convolution kernel weights due to its simplicity, efficiency, and applicability to large-scale datasets~\cite{r87,r88,r89,r90,r91,r92,r93}. Its learning rule is as:
\begin{equation}\label{sgd}
	{\theta ^{t + 1}} = {\theta ^t} - \eta {\nabla _\theta }\left( {{\theta ^t}} \right),
\end{equation}
where $ \theta $ denotes the model parameters, $ t $ denotes the iteration round, and $ \eta $ denotes the learning rate. Therefore, the stochastic gradients of the model parameters (The derivations of $ s_{ir} $ and $ a_i $ apply to $ u_{jr} $, $ e_j $ and $ o_k $, respectively.) are given as:
\begin{equation}
	\left\{ \begin{array}{l}
		\frac{{\partial {\varepsilon _{ijk}}}}{{\partial {s_{ir}}}} = {\varphi _{ijk}}\left( {{u_{jr}}{{\tilde v}_{kr}}} \right) + \lambda {s_{ir}}\\
		\frac{{\partial {\varepsilon _{ijk}}}}{{\partial {v_{kr}}}} = {\varphi _{ijk}}\left( {{s_{ir}}{u_{jr}}{{\tilde v}_{kr}}\left( {1 - {{\tilde v}_{kr}}} \right){w_{1r}}} \right) + \lambda {v_{kr}}\\
		\frac{{\partial {\varepsilon _{ijk}}}}{{\partial {w_{cr}}}} = {\varphi _{ijk}}\left( {{s_{ir}}{u_{jr}}{{\tilde v}_{kr}}\left( {1 - {{\tilde v}_{kr}}} \right){v_{\left( {k - c + 1} \right)r}}} \right) + \lambda {w_{cr}}\\
		\frac{{\partial {\varepsilon _{ijk}}}}{{\partial {a_i}}} = {\varphi _{ijk}} + \lambda {a_i}
	\end{array} \right.
\end{equation}
where $ \varphi _{ijk} = - \left( {{x_{ijk}} - {{\tilde x}_{ijk}}} \right){\tilde x_{ijk}}\left( {1 - {{\tilde x}_{ijk}}} \right) $. With (\ref{sgd}), their update rules are as:
\begin{equation}
	\left\{ \begin{array}{l}
		\!	\!	s_{ir}^{t + 1} \!=\! s_{ir}^t \!-\! \eta \left( {{\varphi _{ijk}}\left( {{u_{jr}}{{\tilde v}_{kr}}} \right) \!+\! \lambda s_{ir}^t} \right)\\
		\!	\!	v_{kr}^{t + 1}\! =\! v_{kr}^t \!-\! \eta \left( {{\varphi _{ijk}}\left( {{s_{ir}}{u_{jr}}{{\tilde v}_{kr}}\left( {1 \!-\! {{\tilde v}_{kr}}} \right){w_{1r}}} \right)\! +\! \lambda v_{kr}^t} \right)\\
		\!	\!	w_{cr}^{t + 1}\! = \!w_{cr}^t \!-\! \eta \left( {{\varphi _{ijk}}\left( {{s_{ir}}{u_{jr}}{{\tilde v}_{kr}}\left( {1 \!-\! {{\tilde v}_{kr}}} \right){v_{\left( {k - c + 1} \right)r}}} \right) \!\!+\!\! \lambda w_{cr}^t} \right)\\
		\!	\!	a_i^{t + 1}\! = \!a_i^t \!-\! \eta \left( {{\varphi _{ijk}} \!+\! \lambda a_i^t} \right)
	\end{array} \right.
\end{equation}

\subsection{The PSO-based Hyperparameters Adaptation}
The learning of model parameters relies on two hyperparameters $ \eta $ and $ \lambda $. We adopt a PSO-based hyperparameters adaptation scheme to reduce the large time overhead suffered by manually tuning. Specifically, $ P $ particles are initialized, where the velocity and position of the $ p $-th particle are defined by two vectors, $ \textbf{\textit{m}}_{(p)}=(\eta_{(p)}, \lambda_{(p)}) $ and $ \textbf{\textit{n}}_{(p)}=(n_{\eta(p)}, n_{\lambda(p)}) $, $ p=\{1,2,...,P\} $. Following the PSO principle, $ p$-th particle is updated as:
\begin{equation}
	\left\{ \begin{array}{l}
		\textbf{\textit{n}}_{(p)}^{t + 1} = \omega \textbf{\textit{n}}_{(p)}^t + {c_1}{r_1}\left( {\textbf{\textit{pb}}_{(p)}^t \!-\! \textbf{\textit{m}}_{(p)}^t} \right) \!+\! {c_2}{r_2}\left( {\textbf{\textit{gb}}^{t} - \textbf{\textit{m}}_{(p)}^t} \right),\\
		\textbf{\textit{m}}_{(p)}^{t + 1} = \textbf{\textit{m}}_{(p)}^t + \textbf{\textit{n}}_{(p)}^{t + 1},
	\end{array} \right.
\end{equation}
where $ \omega $ denotes the inertia constant, $ c_1 $ and $ c_2 $ denote the factors, $ r_1 $ and $ r_2 $ denote the random values in the scale of $ [0,1] $, $ \textbf{\textit{pb}}_{(p)} $ and $ \textbf{\textit{gb}} $ denote individual and global optimums, respectively. To evaluate the performance gain of each particle, we adopt the fitness function as:
\begin{equation}
	F_{(p)}^{t+1} = \frac{{Q\left( {\textbf{\textit{m}}_{(p)}^{t+1}} \right) - Q\left( {\textbf{\textit{m}}_{(p - 1)}^{t+1}} \right)}}{{Q\left( {\textbf{\textit{m}}_{(P)}^{t+1}} \right) - Q\left( {\textbf{\textit{m}}_{(P)}^{t}} \right)}},
\end{equation}
where $ Q(\textbf{\textit{m}}^{t+1}_{(0)}) = Q(\textbf{\textit{m}}^{(t)}_{P}) $ and $ Q $ is defined as:
\begin{equation}
	Q\left( {{\textbf{\textit{m}}_{(p)}}} \right) = \sqrt {{{\sum\limits_{{x_{ijk}} \in \Lambda } {{{\left( {{x_{ijk}} - {{\tilde x}_{ijk\left( p \right)}}} \right)}^2}} } \mathord{\left/
				{\vphantom {{\sum\limits_{{x_{ijk}} \in \Lambda } {{{\left( {{x_{ijk}} - {{\tilde x}_{ijk\left( p \right)}}} \right)}^2}} } {\left| \Lambda  \right|}}} \right.
				\kern-\nulldelimiterspace} {\left| \Lambda  \right|}}},
\end{equation}
where $ \Lambda $ denotes the validation set from the unobserved element set. Therefore, the update rules of model parameters with the hyperparameters adaptation are given as (Taking $ s_{ir} $ for example):
\begin{equation}\label{e17}
	s_{ir\left( p \right)}^{t + 1} = s_{ir\left( p \right)}^t - {\eta _{\left( p \right)}}\left( {{\varphi _{ijk}}\left( {{u_{jr}}{{\tilde v}_{kr}}} \right) + {\lambda _{\left( p \right)}}s_{ir\left( p \right)}^t} \right).
\end{equation}

With (\ref{e17}), the proposed CLR algorithm implements the learning of model parameters and automatically updates of hyperparameters. The specific process of the model is presented in Algorithm~\ref{Alg1}.

\begin{algorithm}[t]
	\caption{CLR model}
	\label{Alg1}
	Initialize $\Omega$, $I$, $J$, $K$, $R$, $ \textbf{\textit{m}} $, $ \textbf{\textit{n}} $, $ P $, $ T $ \;	
	Initialize $ \mathrm{S} $, $ \mathrm{U} $, $ \mathrm{V} $, $ \textbf{\textit{a}} $, $ \textbf{\textit{e}} $, $ \textbf{\textit{o}} $, $ \mathrm{W} $ with positive values\;
	\While{$ t \le T$ \rm{and not converge}}
	{
		\ForEach{$ p = 1 $ to $ P $}
		{
			Update model parameters $ \theta $ \;
		}
		Compute $ Q(\textbf{\textit{m}}^{t}_{(p)}) $\;
		\ForEach{$ p = 1 $ to $ P $}
		{
			Compute $ F(\textbf{\textit{m}}^t_{(p)}) $\;
			update $ \textbf{\textit{pb}}_{(p)}^{t} $ and $ \textbf{\textit{gb}}^{t} $\;
		}
		$ t = t + 1 $\;
	}
\end{algorithm}

\section{Experimental comparisons}\label{comparisons}
\subsection{Datasets}
In this paper, we evaluate the performance of the proposed model on three real-world water quality datasets. They are derived from 24 sampling stations of the Hong Kong Environmental Protection Department in Victoria Harbour for 24 water quality parameters over 903 time slots, where D1 contains 129406 surface observed WQD, D2 contains 121218 middle depth observed WQD, and D3 contains 129415 bottom observed WQD. Further, the datasets are divided into the training set $ (\Gamma) $, the validation set $ (\Lambda) $, and the test set $ (\Phi) $ with 1:2:7. For all experiments, all tested models are run twenty times to provide unbiased average results.

\subsection{Evaluation metrics}
In order to fairly measure the performance of the models, the prevalent Root Mean Squared Error (RMSE) and Mean Absolute Error (MAE)~\cite{r66,r67,r68,r71,r72,r73} are adopted for evaluating the metrics as:
\begin{equation}
	\begin{array}{l}
		{\rm{RMSE}} = \sqrt {{{\sum\limits_{{x_{ijk}} \in \Phi } {{{\left( {{x_{ijk}} - {{\tilde x}_{ijk}}} \right)}^2}} } \mathord{\left/
					{\vphantom {{\sum\limits_{{x_{ijk}} \in \Phi } {{{\left( {{x_{ijk}} - {{\tilde x}_{ijk}}} \right)}^2}} } {\left| \Phi  \right|}}} \right.
					\kern-\nulldelimiterspace} {\left| \Phi  \right|}}} ;\\
		{\rm{MAE}} = {{\sum\limits_{{x_{ijk}} \in \Phi } {{{\left| {{x_{ijk}} - {{\tilde x}_{ijk}}} \right|}_{{\rm{abs}}}}} } \mathord{\left/
				{\vphantom {{\sum\limits_{{x_{ijk}} \in \Phi } {{{\left| {{x_{ijk}} - {{\tilde x}_{ijk}}} \right|}_{{\rm{abs}}}}} } {\left| \Phi  \right|}}} \right.
				\kern-\nulldelimiterspace} {\left| \Phi  \right|}},
	\end{array}
\end{equation}
where $ |\cdot|_{\rm abs} $ denotes the absolute value. Note smaller RMSE and MAE indicate better imputation accuracy.

\subsection{Comparison Results}

We compare the results of the proposed CLR model with four state-of-the-art imputation models on three datasets in terms of imputation accuracy and time cost. The tested models are: a) M1: The proposed CLR model; b) M2~\cite{r6}: A biased low-rank representation model for imputation that learns model parameters via the nonnegative update algorithm; c) M3~\cite{ye2021outlier}: A robust imputation model employing the Cauchy loss function; d) M4~\cite{wang2016multi}: A multilinear imputation model with integrated reconfiguration optimization algorithm; e) M5~\cite{su2021tensor}: A multidimensional imputation model employing the alternating least squares algorithm.

\begin{table}[t]
	\caption{The imputation accuracy for five tested models}
	\begin{center}
		\begin{tabular}{ccccccc}
			\hline
			\textbf{Dataset} &  & \textbf{M1} & \textbf{M2} & \textbf{M3} & \textbf{M4} & \textbf{M5} \\ \hline
			\multirow{2}{*}{\textbf{D1}} & \textbf{RMSE} & \textbf{0.0246} & 0.0256 & 0.0260 & 0.0273 & 0.0267 \\
			& \textbf{MAE}  & \textbf{0.0131} & 0.0162 & 0.0148 & 0.0162 & 0.0163 \\
			\multirow{2}{*}{\textbf{D2}} & \textbf{RMSE} & \textbf{0.0257} & 0.0262 & 0.0266 & 0.0287 & 0.0270 \\
			& \textbf{MAE}  & \textbf{0.0152} & 0.0175 & 0.0161 & 0.0177 & 0.0176 \\
			\multirow{2}{*}{\textbf{D3}} & \textbf{RMSE} & \textbf{0.0249} & 0.0258 & 0.0262 & 0.0277 & 0.0259 \\
			& \textbf{MAE}  & \textbf{0.0158} & 0.0176 & 0.0159 & 0.0175 & 0.0166 \\\hline
		\end{tabular}
		\label{tp}
	\end{center}
\end{table}

\begin{table}[t]
	\caption{The time cost for five tested models (Secs.)}
	\begin{center}
		\begin{tabular}{ccccccc}
			\hline
			\textbf{Dataset} &  & \textbf{M1} & \textbf{M2} & \textbf{M3} & \textbf{M4} & \textbf{M5} \\ 	\hline
			\multirow{2}{*}{\textbf{D1}} & \textbf{Time-R} & \textbf{1.457} & 3.268 & 5.716 & 20.03 & 4.984 \\
			& \textbf{Time-M} & \textbf{1.311} & 4.297 & 3.576 & 20.14 & 3.971 \\
			\multirow{2}{*}{\textbf{D2}} & \textbf{Time-R} & \textbf{1.922} & 4.518 & 4.823 & 15.71 & 7.324 \\
			& \textbf{Time-M} & \textbf{1.820} & 6.248 & 3.862 & 16.42 & 7.146 \\
			\multirow{2}{*}{\textbf{D3}} & \textbf{Time-R} & \textbf{1.796} & 2.465 & 3.146 & 16.65 & 11.06 \\
			& \textbf{Time-M} & \textbf{1.672} & 3.162 & 4.597 & 17.51 & 11.15 \\ \hline
		\end{tabular}
		\label{tc}
	\end{center}
\end{table}

The rank of all models are set to 10 and the iteration rounds are set to 1000. Note that the tested models are terminated if the error in two consecutive iteration rounds is less than $ 10^{-5} $. Tables~\ref{tp} and~\ref{tc} present the statistical results, and it can be observed that:
\begin{itemize}
	\item \textit{The CLR model presents higher imputation accuracy compared with its peers.} As shown in Table~\ref{tp}, the RMSE and MAE on D2 of M1 are 0.0257 and 0.0152, respectively. It improves by 1.90\% and 13.14\%, 3.38\% and 5.59\%, 10.45\% and 14.12\%, 4.48\% and 13.63\% compared to 0.0262 and 0.0175, 0.0266 and 0.0161, 0.0287 and 0.0177, 0.0270 and 0.0176 of M2-M5, respectively. Similar performance comparisons are also observed on other datasets.
	\item \textit{The CLR model exhibits less time cost compared with its peers.} For the statistical results in Table~\ref{tc}, M1 costs 1.922 seconds in RMSE  and 1.820 seconds in MAE on D2. Compared to M2's 4.518 and 6.248, M3's 4.823 and 3.862, M4's 15.71 and 16.42, M5's 7.324 and 7.146, M1's time cost in RMSE and MAE are only their 42.54\% and 29.12\%, 39.85\% and 47.12\%, 12.23\% and 11.08\%, 26.24\% and 25.46\%, respectively. On the other two datasets, the time cost of M1 are similarly less than that of the comparison models.
\end{itemize}

Based on the above analysis, the proposed CLR model achieves higher imputation accuracy and less time cost against to the other imputation models.

\section{Conclusions}\label{conclusions}
To achieve precise imputation of missing data, this paper proposes a low-rank representation model with higher imputation accuracy and less time cost. Specifically, it employs a causal convolution operator to model temporal features and fuse temporal information. In addition, it achieves efficient model training via an adaptation scheme. Model performance evaluation on three water quality datasets indicated that the proposed model outperforms existing imputation models for imputing the missing data. For future work, we plan to explore the low-rank representation of spatial information~\cite{r74,r75,r76,r11}.

\bibliographystyle{IEEEtran}
\bibliography{CLR}

\vspace{12pt}

\end{document}